# New Hoopoe Heuristic Optimization

Mohammed El-Dosuky[1], Ahmed EL-Bassiouny[2], Taher Hamza[3] and Magdy Rashad[4]

[1] Computer sciences Department, Faculty of Computers and Information,
Mansoura University, Egypt
*mouh_sal_010@mans.edu.eg*

[2] Department of Mathematics, Faculty of Sciences
Mansoura University, Egypt
*el_bassiouny@mans.edu.eg*

[3] Computer sciences Department, Faculty of Computers and Information,
Mansoura University, Egypt
*Taher_Hamza@yahoo.com*

[4] Computer sciences Department, Faculty of Computers and Information,
Mansoura University, Egypt
*magdi_12003@yahoo.com*

*Abstract*— Most optimization problems in real life applications are often highly nonlinear. Local optimization algorithms do not give the desired performance. So, only global optimization algorithms should be used to obtain optimal solutions. This paper introduces a new nature-inspired metaheuristic optimization algorithm, called Hoopoe Heuristic (HH). In this paper, we will study HH and validate it against some test functions. Investigations show that it is very promising and could be seen as an optimization of the powerful algorithm of cuckoo search. Finally, we discuss the features of Hoopoe Heuristic and propose topics for further studies.

I. INTRODUCTION

Most optimization problems in real life applications are often highly nonlinear. Local optimization algorithms do not give the desired performance. So, only global optimization algorithms should be used to obtain optimal solutions ([2] [9]).

It is Glover who first mentioned the term *metaheuristic*, when he proposed the *tabu search* [14]. Metaheuristics refer to a class of global heuristic optimizers. Many nature-inspired metaheuristic optimization algorithms are proposed to imitate the best behaviors in nature [34]. Farmer et al. contribute the *artificial immune system (AIS)* [11]. Goldberg contributes genetic algorithms [16]. Dorigo contributes the ant colony optimization (ACO) in his PhD thesis [10]. Kennedy and Eberhart propose *particle swarm optimization (PSO)* [21]. Karaboga contributes *Artificial Bee Colony Algorithm (ABC)* [20]. Yang and Deb propose *cuckoo search* [35].

When a metaheuristic explores a search space, it has two components of *intensification* and *diversification* ([7], [13]). These two strategies originate back to the *tabu search*, where intensification focuses on examining neighbors of elite solutions and diversification encourages examining unvisited regions [15]. Intensification is a deterministic component and diversification is a stochastic component [19]. Metaheuristic algorithms should be designed so that intensification and diversification play balanced roles [7]. Algorithm shown in Fig. 1 is a rough algorithmic skeleton on how a metaheuristic algorithm works [36].

```
Create one or several start solutions, randomly
    while termination criterion not satisfied do
        if intensify then
            Create new solution by intensification;
        else
            Create new solution by diversification;
        end
        Update best found solution (if necessary);
    end
return Best found solution;
```

Figure 1. Abstract algorithmic framework for metaheuristics [36]

This abstract algorithm unifies our conceptualization of metaheuristics. But it does not show when to actually perform intensification or diversification. Additionally it does not show how to terminate the search.

We develop a new metaheuristic search algorithm called Hoopoe Heuristic (HH). In this paper, we will study HH and validate it against some test functions. Investigations show that it is very promising and could be seen as an optimization of the powerful algorithm of *cuckoo search*. Finally, we will discuss the unique features of Hoopoe Heuristic and propose topics for further studies.

## II. HOOPOE HEURISTIC

Let us first review the interesting ground probing behavior of certain hoopoe species, to facilitate describing the proposed Hoopoe Heuristic.

### A. Hoopoe Ground Probing Behavior

Hoopoe is an interesting bird. There is a similarity between the Hoopoe and the woodhoopoes. Hoopoe is a solitary forager feeding on worms and seeds that are usually hidden beneath the surface. Hoopoes forage mainly in short grass and on bare soil. Hoopoe constantly makes short probes into the ground and sometimes pausing to insert the bill fully. The detected eatable objects are extracted or dug out. ([12], [17], [22])

A similar probing and subsurface investigation mechanism is used in geophysics, either direct or indirect methods. Direct methods apply boring and sampling to measure in situ properties, performing too much testing is an issue. Indirect methods apply remote sensing techniques ([23], [24])

Common direct site investigation methods include: Standard Penetration Test (SPT) using a standard split barrel sampler, Cone Penetration Test (CPT) measuring in situ shear strength of soil, Vane Shear Test based on measuring shear strength for soft clay, Dilatometer Test (DMT) for lateral stresses, and Pressuremeter Test (PMT) for both shear strength and lateral pressures. Common indirect site investigation methods include: Resistivity by measuring the resistance to current flow, Seismic Waves be transmitted and their measuring their velocities, Magnetometry used to pick up magnetic anomalies in the subsurface, and Ground Penetrating Radar (GRP) which are radio frequency microwaves capable of penetrating the soil ([23], [24]) .

We prefer the direct Cone Penetration Test (CPT), by testing the neighbors in circle whose center is the current probing point, with a fixed radius or made to depend on region area. Clearly, Ground Probing is an intensification step.

### B. Lévy-flight foraging

Predators search for preys by exploring their landscape. The foraging path is a random walk because the next move decision depends on the current location and the transition probability to the next location [25]. Many recent studies assure two patterns of random walks. They found that Brownian movement is associated with abundant preys and Lévy flight is associated with sparser or unpredictably distributed preys ([5], [6], [31]).

With apparent potential capability, Lévy flight is applied to optimal search [25].

Lévy flight is applied to generate new states $s^{(t+1)}$ based on current state.

$$s^{(t+1)} = s^{(t)} + \alpha * \text{Lévy}(\lambda) \tag{1}$$

where $\alpha > 0$ is the step size, we can use $\alpha = O(1)$. The product means entry-wise multiplications. Lévy flights obey a Lévy distribution for large steps

$$\text{Lévy} \sim u = t^{-\lambda}, \quad (1 < \lambda \leq 3) \tag{2}$$

which has an infinite variance with an infinite mean. But for simplicity, jumps of a hoopoe can be made to obey a step-length distribution with a heavy tail. Clearly, Lévy flight is diversification step.

### C. Hoopoe Heuristic

Let us first state the following general assumptions:

- Only one hoopoe is needed to explore the landscape randomly using Lévy-flight path;
- The hoopoe probes the current region by penetrating it with its bill, with a success probability to find a worm, $p_s \in [0, 1]$.
- The hoopoe digs a region with high quality of probe, i.e., that has a success probability greater than a predefined threshold. This means to explore the neighborhood of the penetration point.

Based on these assumptions, the basic steps of the Hoopoe Heuristic are shown in Fig. 2.

```
function HOOPOE_HEURISTIC (pop, FIT-FN, θ)
returns a point
input:
pop, population, a set of points (xi, yi)
FIT-FN, a function telling the quality of a point
θ, intensification threshold θ∈[0..1]
Variables:
current, current region, a subset of population
closed, a set to contain all explored regions
r, preferred digging radius, fixed value
best, best solution
best_fitness, fitness of best solution
Begin
    best ← NULL
    best_fitness ← 0
    closed ← { }
    current ← RANDOM_SELECT(pop)
    while termination_criterion not satisfied do
        closed ← closed ∪ current
        if |closed|/|pop| > then
            current ← Ground-Probing(current, pop, r)
        else
            current ← Lévy-Flight (current, pop)
        end if
        Rank current using FIT-FN
        if MaxFitness(current) > best_fitness then
            best_fitness ← MaxFitness(current)
            best ← FIND (current, MaxFit(current) )
        end if;
    end while;
return best
```

Figure 2. Hoopoe Heuristic

## III. VALIDATION

We review some test functions. Then we compare the proposed algorithm with a common previous work.

### A. Standard Test functions

There are many benchmark test functions to test the performance of optimization algorithms, though there is no unified standard list ([28], [29]). There are certain guidelines for selecting the test function. The most important criteria are to contain nonlinear non-separable problems and to include high-dimensional functions with a large number of local optima ([4], [33]). Our test framework contains the following four functions, shown in Fig.3.

De Jong function [8]

$$f_1(x) = \sum_{i=1}^{n} x_i^2 \tag{4}$$

Rosenbrock function [27]

$$f_2(x) = \sum_{i=1}^{n-1}(100(x_i^2 - x_{i+1})^2 + (1-x_i)^2) \quad (5)$$

The generalized Ackley function ([1], [3])

$$f_3(x) = -20\exp\left(-0.2\sqrt{\frac{1}{n}\sum_{i=1}^{n}x_i^2}\right) - \exp\left(\frac{1}{n}\sum_{i=1}^{n}\cos(2\pi x_i)\right) + 20 + e \quad (6)$$

The generalized Rastrigin function [18]

$$f_4(x) = 10n + \sum_{i=1}^{n}\left(x_i^2 - 10\cos(2\pi x_i)\right) \quad (7)$$

## B. Results

We conduct 100 simulation runs of the algorithm and Matlab implementation of Cuckoo Search ([26], [32]). For the four test functions described in section 3.1: De Jong (32 dimension), Rosenrbrock(16 dimension), Ackley (128 dimension), (6) Rastrigin (default dimensions). Table 1 shows standard deviation and its percentage success rate of finding the global optima, and Fig. 4 shows this graphically.

TABLE I. COMPARISON OF HOOPOE HEURISTIC WITH CUCKOO SEARCH

| Functions | cuckoo | Hoopoe |
|---|---|---|
| De Jong | 540(100%) | 537(100%) |
| Rosenrbrock | 1938(100%) | 1932(100%) |
| Ackley | 907(100%) | 897(100%) |
| Rastrigin | 3748(100%) | 3648(100%) |

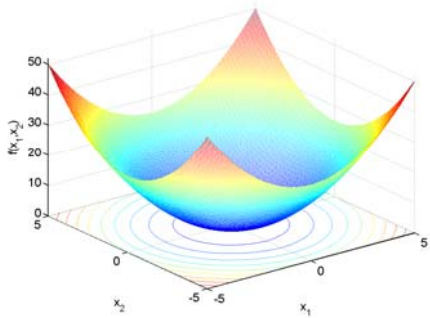

a) De Jong function

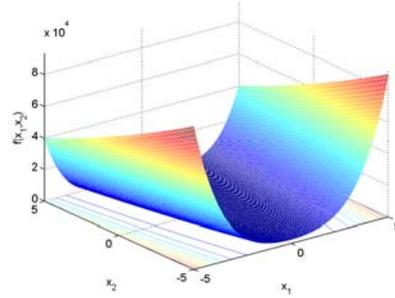

b) Rosenbrock function

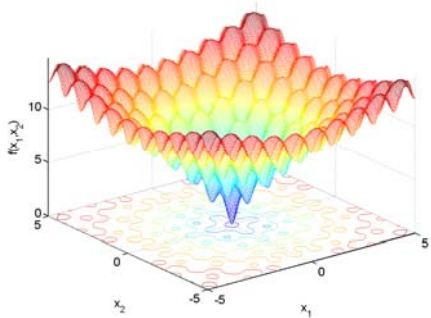

c) Ackley function

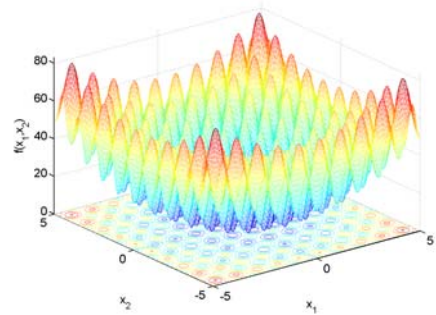

d) Rastrigin function

Figure 3. Standard Test functions

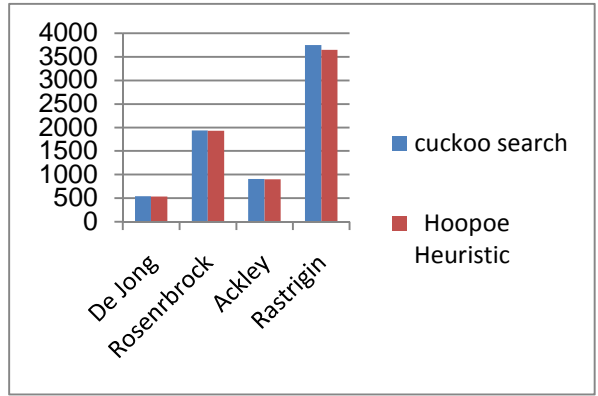

Figure 4. Comparison of Hoopoe Heuristic with cuckoo search

*C. Comparison*

The Hoopoe has an affinity with cuckoo! There are many similarities between the two algorithms that amazed us. An egg of a cuckoo to put corresponds to a worm of a hoopoe to dig for. The success probability $P_s$ of a hoopoe is the complementary of the failure probability $P_a$ of a cuckoo.

There are three major differences between the two algorithms are:

- The Hoopoe Heuristic explicitly distinguishes the two steps of intensification and diversification;

- The Hoopoe Heuristic utilizes a unique step of Ground-Probing, which change the philosophy from generate-and-test to test-to-generate. This leads to the third difference that;

The Hoopoe Heuristic uses the success probability as an intensification measure. With an intensification threshold, the algorithm can stop earlier. Fig. 5 indicates the stopping moment for both algorithms.

IV. DISCUSSIONS AND CONCLUSIONS

We have introduced a new nature-inspired metaheuristic optimization algorithm, called Hoopoe Heuristic (HH). The algorithm generates random moves using Lévy-Flight. After a while, the Hoopoe Heuristic changes its behavior by generating random moves using Ground-Probing. This may be interpreted by the level of experience a hoopoe has after a time spent in investigating the landscape.

After validating the Hoopoe Heuristic against some test functions, investigations show that it is very promising.

Future work can be divided along many directions. First, investigating the use other Ground-Probing heuristics described in section 2.1, by applying equations of remote sensing methods such as Resistivity, Magnetometry, and Ground Penetrating Radar (GRP). These methods are non-impact methods that may help handling massive data.

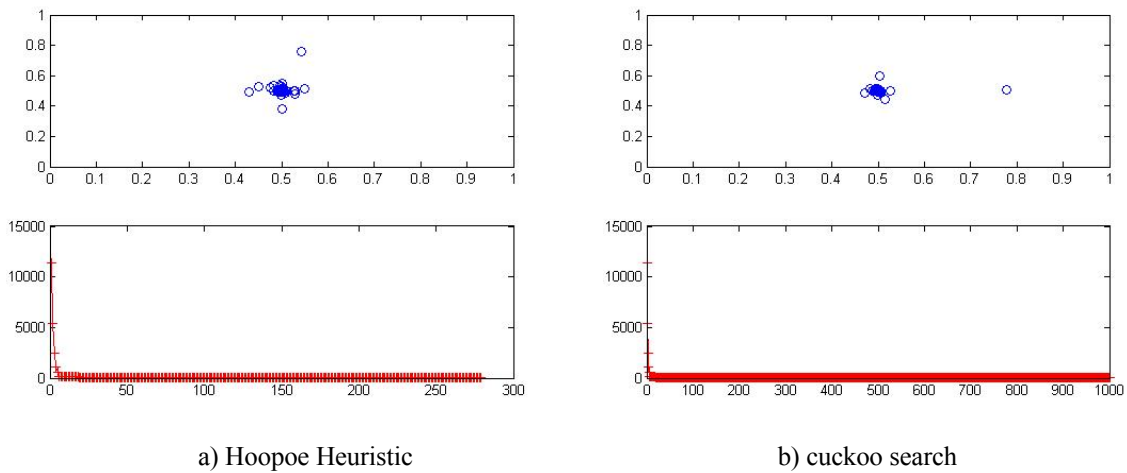

a) Hoopoe Heuristic  b) cuckoo search

Figure 5. Hoopoe Heuristic stooping earlier than cuckoo search

Another future direction is investigating the concept of intensification measures. Finally, an important future direction is applying the algorithm to complex real life problems such as mining social networks.